\pdfoutput=1
\documentclass[10pt, conference, compsocconf]{IEEEtran}
\ifCLASSINFOpdf
\else
\fi
\usepackage{amsmath}
\usepackage{booktabs}
\usepackage{pifont}
\usepackage{url}
\usepackage{amsfonts}
\usepackage{amssymb}
\usepackage{mathtools}
\usepackage{multirow}
\usepackage{hyperref}
\hypersetup{colorlinks=true, linkcolor=blue}
\usepackage{float}

\usepackage{graphicx}

\begin{document}

\title{STF: Spatio-Temporal Fusion Module for Improving Video Object Detection }


\author{
  \IEEEauthorblockN{Noreen Anwar, Guillaume-Alexandre Bilodeau}
  \IEEEauthorblockA{LITIV, Polytechnique Montr\'eal,\\
    Montr\'eal, Canada\\
    \{noreen. anwar, gabilodeau\}@polymtl.ca}
    
  \and
  
  \IEEEauthorblockN{Wassim Bouachir}
  \IEEEauthorblockA{Data Science Laboratory, \\
  University of Quebec (TELUQ)\\
    Montr\'eal, Canada\\
    wassim.bouachir@teluq.ca}
}


%Telephone: (800) 555--1212, Fax: (888) 555--1212}


\maketitle

\begin{abstract}
Consecutive frames in a video contain redundancy, but they may also contain relevant complementary information for the detection task. The objective of our work is to leverage this complementary information to improve detection. Therefore, we propose a spatio-temporal fusion framework (STF). We first introduce multi-frame and single-frame attention modules that allow a neural network to share feature maps between nearby frames to obtain more robust object representations. Second, we introduce a dual-frame fusion module that merges feature maps in a learnable manner to improve them.
Our evaluation is conducted on three different benchmarks including video sequences of moving road users. The performed experiments demonstrate that the proposed spatio-temporal fusion module leads to improved detection performance compared to baseline object detectors. Code is available at \url{https://github.com/noreenanwar/STF-module}
\end{abstract}


% \fi
\IEEEpeerreviewmaketitle

\section{Introduction}
Computer vision has made remarkable progress in object detection from a single frame for localizing and identifying objects (\cite{tian2019fcos,lin2017feature}). However, relying solely on a single frame for detection is not always effective and sufficient as argued in several recent works (\cite{zhu2017flow,zhou2020tracking,liu2018mobile,perreault2021ffavod,broad2018recurrent,bertasius2018object}). Single-frame object detectors are subject to errors in the case of poor or improper visibility of objects that can be caused by occlusions, motion blur, or small object sizes. When objects are occluded or in the case of motion blur, their appearance features can be severely altered. The object detector should be robust to the fact that objects can exist on a spectrum of scales and sizes. Furthermore, small objects have less distinctive features making them harder to detect. 

To address these problems, as in some previous work (\cite{zhu2017flow,zhou2020tracking,liu2018mobile,perreault2021ffavod,broad2018recurrent,bertasius2018object,kang2017t,lee2016multi}), we propose to use multiple frames for better features representation. Given that we are interested in processing videos for road safety analysis, our work is in a context that fits well with object detection from multiple frames. Having several sequential frames to detect objects has the significant benefit of providing temporal complementary information about a given instance, generally observed over a short time. This kind of temporal information is utilized in some existing multiple-frame object detection methods by first applying a single-frame object detector and then integrating their bounding boxes across frames using an off-the-shelf motion estimation method (\cite{kang2017t,lee2016multi}). The performance improvement relies on heuristic post-processing and those methods do not capitalize on combining features from several frames to compensate for poor feature quality in some frames.

Another solution for multiple-frame detection is to fuse features from several frames together in a learnable manner for better feature alignment (\cite{zhu2017flow,perreault2021ffavod,zhou2020tracking,liu2018mobile,broad2018recurrent}). 
Using multiple frames is not trivial, however as the features of consecutive frames are not always aligned or corresponding to the visibility state of an object that can change (e.g. not the same part of an object might occluded). This means that there are no trivial ways to determine which features are more important for the detection. Therefore, feature fusion has to be done carefully.

Global contextual attention involves capturing long-range dependencies and relationships between different regions of an image to understand specific parts and the context. Therefore, we propose a global contextual attention model for feature selection and fusion from a pair of frames. We present an end-to-end framework that learns multiple frame information and fuses it without prior knowledge of motion or temporal relations. We aim to improve the detection accuracy by effectively utilizing temporal and spatial information from two frames, the current frame and the past frame. As mentioned above, it is important to consider that the features corresponding to the same object instances in two frames often lack spatial alignment across frames due to movements or occlusions. To take this into account, we introduced multi-frame (temporal) and single-frame (spatial) attention-based modules. Secondly, to handle small objects, we are considering the multiple layer resolution features from our backbone. Our attention modules operate on those multiple resolution layers. 

Our proposed approach, STF (Spatio-temporal fusion), is based on per-frame feature learning through temporal and spatial fusion of features from the current and a past frame. To achieve this, we are proposing two new attention-based modules: the first applies multi-frame attention, while the second applies single-frame attention. Here, we hypothesize that global contextual information along with spatio-temporal information can address the detection problems better as compared to previous works, limited to single-frames 
and multi-frame methods that fuse feature maps without attention 
(\cite{zhu2017flow,zhou2020tracking,zhu2019empirical}). Then, our dual-frame fusion module helps to fuse the learned features from the past and current frames to improve detection accuracy under challenging conditions, like occlusion or motion blur. The effectiveness of our method is evaluated on three popular traffic-related datasets, including KITTI MOT~\cite{geiger2012we}, Cityscapes \cite{elsayed2020revisiting} and UAVDT \cite{du2018unmanned} and we obtained competitive results compared to SOTA detectors.

Our main contribution is the introduction of an end-to-end learnable fusion module that combines the current and a past frame by utilizing their temporal, spatial and channel features information. Our specific contributions can be outlined as:

\begin{itemize}
    \item 
A multi-frame attention (MFA) module with temporal convolutions used after the backbone feature extractor to efficiently use the feature maps of two frames, and enhance features for detecting occluded or blurred objects;
    \item 
A single-frame attention (SFA) module that weights the current frame feature maps in channel and spatial dimensions to reduce false positive detection;

\item An efficient dual-frame fusion module to integrate single-frame and multi-frame feature maps at different scales.

%\item Our model outperforms the state-of-the-art (SOTA) models for sequential frames over multiple datasets.
    
\end{itemize}

\section{Related work}
Using multiple frames in object detection was studied in several previous works because it facilitates the association of detected objects, thereby improving the precision and resilience of the detection process. It consists of detecting objects in frames using their spatial and temporal features. Nevertheless, the study of video-based object detection is receiving comparatively less attention than single-frame detectors, yet their applications are numerous and impactful, which includes video surveillance for security, robot navigation, and autonomous driving. Sequential frames have significant complementary information about the same instances, generally observed in multiple frames during a short period of time. Existing multiple-frame object detection methods, such as those proposed by Kang et al. (\cite{kang2017t}) and Lee et al.~\cite{lee2016multi}, readily capture this type of temporal information. These methods first apply single-frame object detectors and then integrate bounding boxes across frames using off-the-shelf motion estimation, which may compromise the quality of detection due to hand-crafted rules. The improvement in the performance depends on heuristic post-processing through box-level methods without end-to-end training.

Zhu et al. \cite{zhu2017flow} introduced flow-guided feature aggregation, where optical flow warping is used to integrate feature maps from temporally adjacent frames in order to increase detection accuracy. There is another way, proposed in \cite{bertasius2018object} that calculates the offsets between temporally adjacent frames. These offsets enable the sharing of features from adjacent frames, improving the ability to perform the detection tasks.
Similarly, there is another approach, known as FFAVOD (Feature Fusion Architecture for Video Object Detection)~\cite{perreault2021ffavod}, which shares feature maps between nearby frames. FFAVOD proposes a feature fusion module that learns to merge feature maps to improve video-based object detection and classification. RN-VID\cite{perreault2020rn} uses information from nearby frames and merges feature maps of similar dimensions using $1\times 1$ convolution and re-ordering of channels to enhance detection. Zhou et al. \cite{zhou2020tracking} presented CenterTrack, a method that uses a point-based framework to perform simultaneous detection and tracking of objects. This method concatenates two frames and a prior heatmap as input and associates objects through time while performing the detection from the two frames. 
Previous research also explores using both motion and appearance cues of objects in a video sequence with models such as Recurrent Neural Networks (RNNs). Using an RNN, the method named Spatio-Temporal Memory module (STMM) \cite{xiao2018video}  introduced a concatenated spatial-temporal memory across consecutive frames to improve detection. Additionally, Long Short-Term Memories (LSTMs) have been employed to interpolate feature maps, resulting in a notable improvement in inference speed \cite{liu2018mobile}. The Recurrent Multi-frame Single Shot Detector (MF-SSD)  method combines features extracted from multiple consecutive frames~\cite{broad2018recurrent}. This is achieved through the integration of a recurrent convolutional module, enabling the integration of characteristics that extend across multiple frames.

%In the context of combining features from multiple frames, various feature fusion strategies have been explored. Most strategies consider concatenating features or summing them. For instance, the inception network uses convolutions with kernels of different sizes to generate a set of feature maps, which are then combined through concatenation \cite{szegedy2015going}. FSSD adopts a top-down approach, combining pyramid levels through up-sampling and summation \cite{li2017fssd}. In this method, a modified feature pyramid is proposed, wherein feature maps from all levels are initially concatenated and subsequently used to produce all pyramid levels. In the case of ExFuse, a fusion of low-level and high-level semantic information occurs by introducing spatial information into high-level features and semantic information into low-level features~\cite{zhang2018exfuse}. The combination of feature maps in this method is achieved through summation. 

The above mentioned works are mainly focusing on either concatenating or simply summing feature maps rather than using a more fully learnable way. Unlike these methods, our approach aims to train a learnable fusion-based module including temporal, spatial, and channel-based feature information, in a completely end-to-end manner, using the current frame and a past frame.

\section{Methodology}
\subsection{Overview}
\begin{figure*}[ht]
\centering
\includegraphics[width=0.6\textwidth]{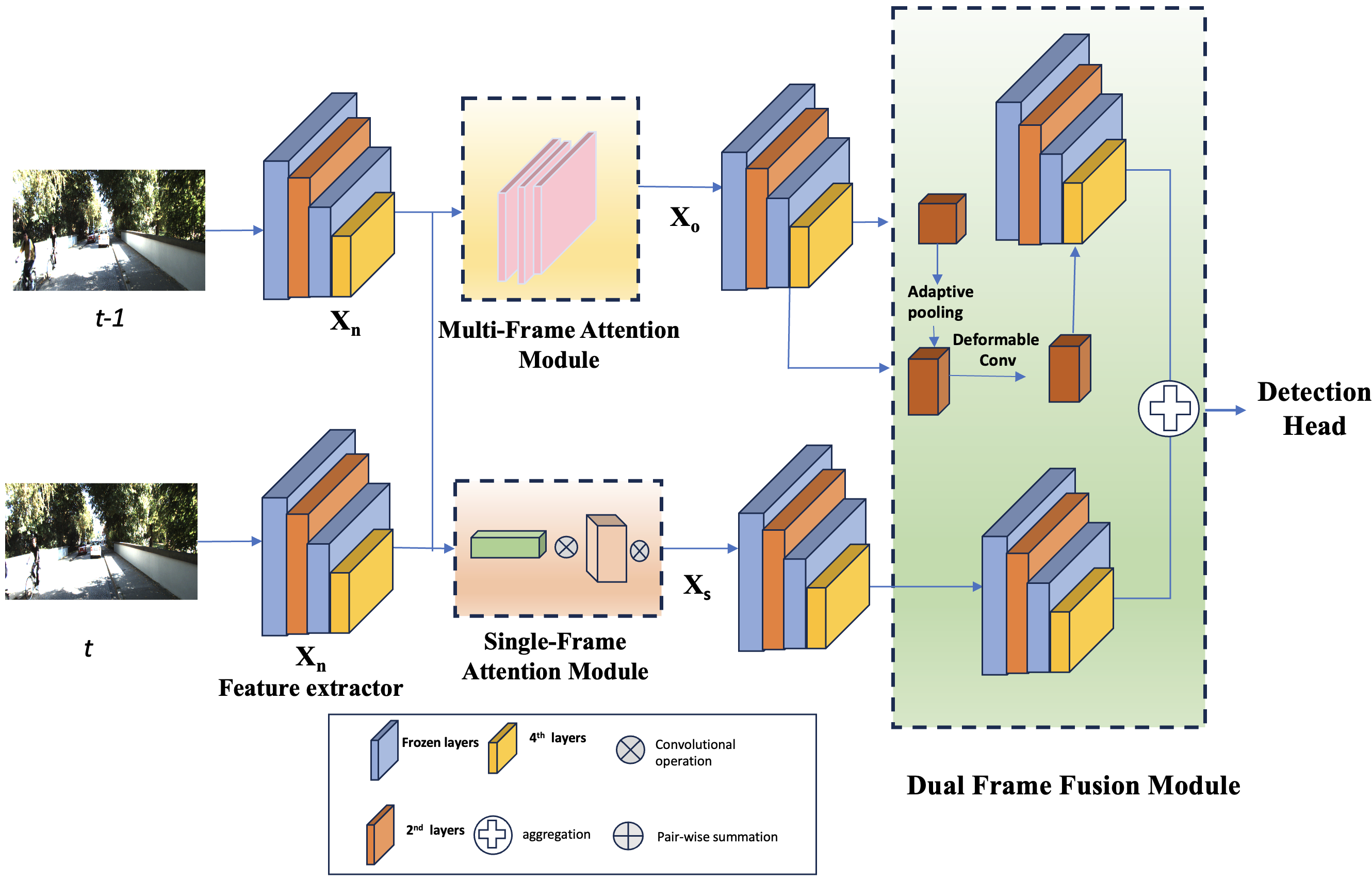} 
\caption{Overview of our spatio-temporal based fusion framework (STF), illustrating the key components: MFA, SFA, and dual-fusion module}
\label{fig:model}
\end{figure*}

The overview of our attention-based framework, STF,  is shown in Figure $\ref{fig:model}$. Given a pair of frames, a pre-trained HRNet \cite{wang2020deep}, where we froze the first and third layers, is used to extract features. After that, the features go through two attention modules: 1) a multi-frame attention (MFA) module that uses the two extracted feature maps to perform temporal and spatial attention, assigning adaptive temporal weights to them, and 2) a single-frame attention (SFA) module that uses spatial and channel dimension attention for improving current frame feature maps. To use the temporally prior frame, the idea here is to combine in a learnable manner the extracted features of the past and current frames for object detection. To combine features from two frames after applying attention, our proposed network fuses temporal, channel, and spatial information by aggregating them at the same time with our dual-frame fusion module. In the following, we introduce these modules in detail.

\subsection{Multi-Frame Attention (MFA) module}

\begin{figure}[ht]
\centering
\includegraphics[width=0.35\textwidth]{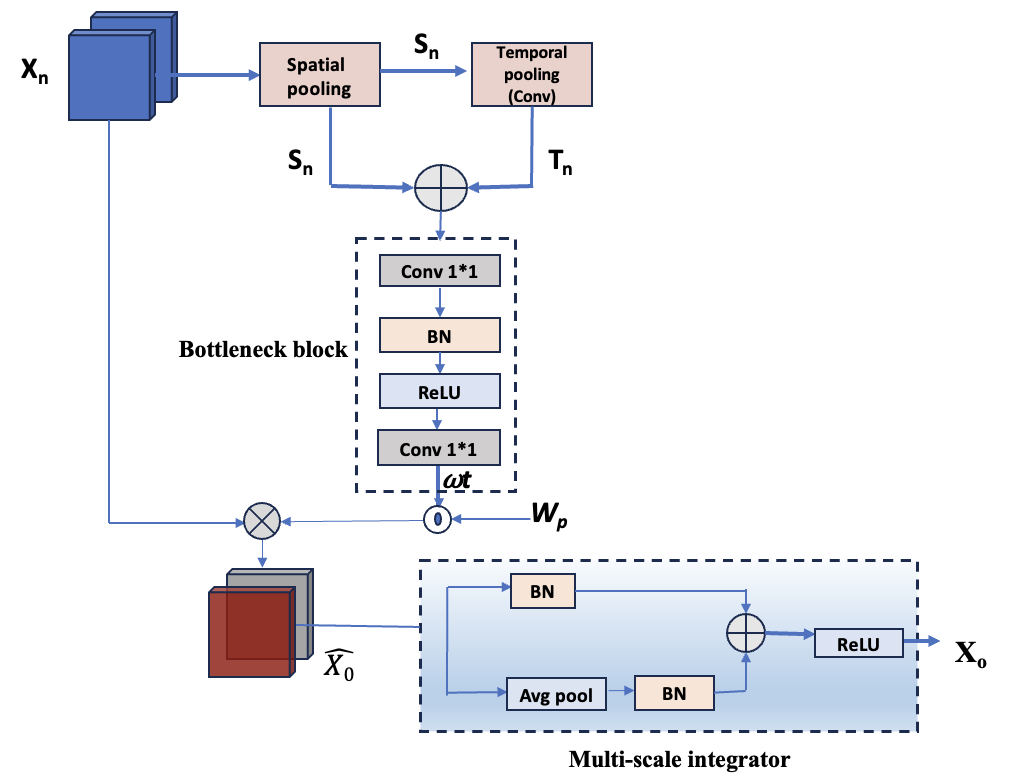}
\caption{Multi-Frame Attention module with multi-scale integrator.}
\label{fig:MDA}
\end{figure}

Given an input video, the multi-scale feature maps of two frames (the current frame and a past frame) are extracted with the HRNet backbone. Then, our goal is to merge the features of these two frames. The Tada Convolution, introduced in the work by Huang et al. \cite{huang2021tada}, efficiently addresses temporal modeling by introducing flexibility to the temporal invariance of 2D convolutions. This is achieved through the incorporation of adaptive temporal weights, which are superimposed onto the convolutional process. Similarly, Cao et al. proposed TCTrack \cite{cao2022tctrack}, which exemplifies the application of Tada Convolution for improving object tracking. This approach employed Tada Convolutions to incorporate adaptive temporal weights, contributing to improved temporal modeling.
Inspired by this previous research, to get adaptive temporal weights for each frame, we designed a Multi-Frame Attention (MFA) module (see Figure \ref{fig:MDA}). The key idea is to adjust the model behavior in real-time as it processes each sequence of frames. This deals with size variations, movement, overlapping, or interaction of objects in frames.

Global information in object detection refers to semantic details that are consistent across frames, helping in identifying objects based on shared characteristics, while local temporal information involves using nearby frames to gather information, such as motion, helping to localize objects, especially in cases of uncertainty about their existence in a specific frame. This module improves the representation ability with multi-frame features by: 1) assigning adaptable weights to each frame to enhance the ability to detect and analyze changes over time, 2) combining both global and local information from multi-frames, and 3) better capturing both detail and broader spatial and temporal information using a multi-scale integrator. 

Our MFA module works as follows. Let us assume that we have an input sequence of frames $I_n$ and we get a sequence $X_n\in \mathbb{R}^{B \times C \times T \times H \times W}$ of features outputted by the HRNet backbone, where $B$ is the batch size, $C$ is the number of channels, $T$ is the temporal dimension, and $H$ and $W$ are the spatial dimensions. For capturing the global spatial context, we start with global average pooling (GAP) across the spatial dimension of the past and current frame features. We then obtain frame descriptor $S_n =\text{GAP}_S(X_n)$, that encompasses global spatial context. To integrate local temporal context effectively, global average pooling across both spatial and temporal dimensions is applied to obtain spatio-temporal descriptor 

\begin{equation}
    T_n = \text{GAP}_{st}(X_n).
\end{equation}

Global spatial context and local temporal information are then aggregated, and this combined information is passed through a bottleneck block (BNB). The output of the bottleneck block results in obtaining local weights $\omega_t$, as illustrated in Figure \ref{fig:MDA}. These weights combine the spatial and temporal descriptors after the bottleneck block with
\begin{equation}
    \omega_t = \text{BNB}(S_n + T_n).
\end{equation}

Then, the total weights that we used in our model are the element-wise product of these weights $\omega_t$ and weights $W_p$ that refer to the initial set of weights in the convolution kernel that is shared across all frames. Note that the local weights $\omega_t$ are set to 0 during initialization, which has the advantage of reducing the training time.
% Convolution operation
An adaptive convolution is then applied to the current frame with
\begin{equation}
    \hat{X}_o = \left(\omega_t \odot W_{p}\right)* \ X_n
\end{equation}
where $\odot$ denotes element-wise multiplication.

To effectively integrate spatio-temporal information and address the limitations in spatial features for a given frame, we finally apply a multi-scale integrator as shown in Figure \ref{fig:MDA}. %The reason for this integrator is to allow richer and more comprehensive representation of spatial-temporal features, to improve the overall performance of the model.
It is expressed as
\begin{equation}
X_o = \lambda(\hat{X}_o) + \gamma(\text{AvgPool}(\hat{X}_o)),
\end{equation}
where $\hat{X}_o$ is the output from the adaptive convolution . The operators $\lambda$ and $\gamma$ represent distinct normalization functions. The goal behind using an average pooling (AP) layer is to enlarge the receptive field to capture a wider range of spatial contexts.

\subsection{Single-Frame Attention Module}

Besides temporal attention, attention in the spatial and channel dimensions also provides a potential enhancement for feature maps derived from single-frame images. In the context of Convolutional Neural Networks (CNNs), the attention mechanism assigns an additional weight to individual pixels in a specific dimension, indicating the significance of particular information. These learned weights strengthen valuable features and weaken less useful ones, facilitating feature screening and enhancement. Furthermore, in videos with generally stable backgrounds, spatial and channel attention, as explained by the methodology proposed in Hou et al. \cite{hou2021coordinate}, can efficiently suppress false positive detection in the background area. 

\begin{figure}[ht]
\centering
\includegraphics[width=0.37\textwidth]{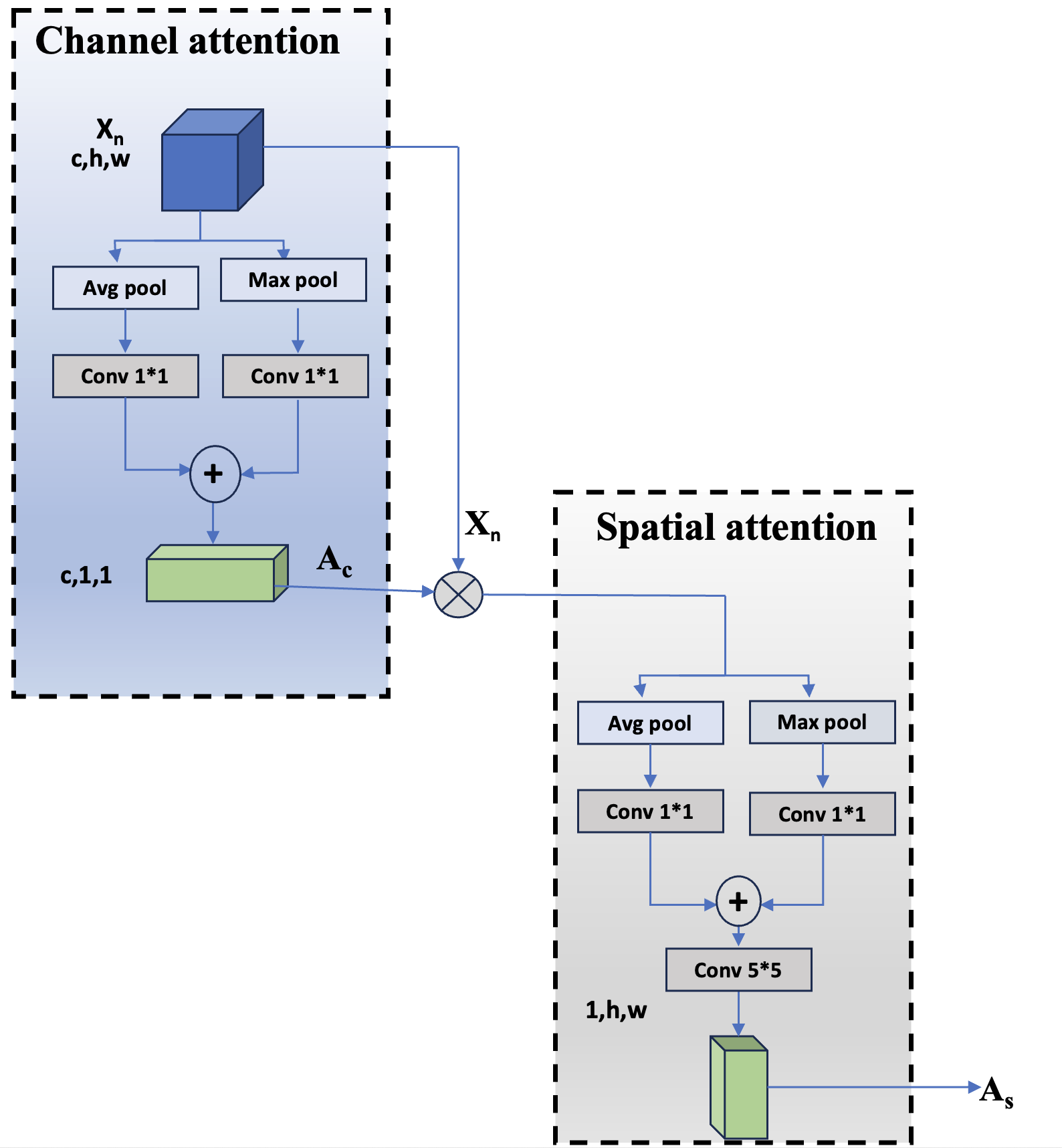}
\caption{The channel and spatial attention modules of our proposed single-frame attention module}
\label{fig:UCAS}
\end{figure}

Inspired by this work, we propose a Single-Frame Attention module (SFA) that uses channel and spatial attention mechanisms, as illustrated in Figure \ref{fig:UCAS}. The SFA module aims to refine feature representation within a single frame. In the SFA module, each frame denoted as $I_n$, is processed to enhance the channel and spatial information of its feature maps $X_n$. First, channel attention with average pooling ($AP$) and max pooling ($MP$) are applied to condense the spatial information. To help our model learn complex feature representation, we integrate $1\times 1$  convolutional layer as shown in Figure \ref{fig:UCAS}. This results in channel attention $A_c$ formulated as
 \begin{equation}
    A_c = Conv_{1\times 1}(AP(X_n)) + (MP(X_n)).
\end{equation}
%Here, $A_c$ i is the channel-based attention from the feature maps. 
For spatial attention, a comparable approach is applied, but it operates within the spatial domain. Here, the features influenced by channel attention are subjected to average and max pooling operations ($MP$), focusing on spatial features. The resulting features are then concatenated and processed through a $5\times5$ convolutional layer to enhance the spatial aspects of in the frame. This gives spatial attention $A_s$ formulated as:
\begin{equation}
\begin{split}
    A_s &= \text{Conv}_{5\times5}\Bigl(\text{Conv}_{1\times 1}(\text{AP}(A_c*X_n)) \\
    &\quad + (\text{MP}(A_c*X_n))\Bigr)
\end{split}
\end{equation}

where $Conv_{5\times5}$ is the convolution operation using a $5\times5$ filter and $\ast$ symbolizes convolution.

Finally, the two attention tensors are concatenated with $X_n$ to obtain the new features $X_s$. This fusion process allows the model to focus on relevant information captured by the attention mechanisms, enhancing the representation of the feature maps. We observed that by using convolutional layers, the module can more effectively capture and enhance the intricate patterns in the features. This ensures that the model is capturing well the spatial features in each frame.

\subsection{Dual-Frame Fusion Module}
%To merge multi-frame temporal information with single-frame channel and spatial attention, we present a dual-frame fusion module. With our feature extractor, as a result we have multi-scale feature maps for each frame. In the SFA module, we applied channel and spatial attention to each feature map of the current frame. However, in the MFA module, the feature maps of the current frame and its preceding frame are concatenated. To obtain attention-weight, we fed these feature maps into an adaptive convolution layer. 
Figure \ref{fig:model} illustrates the feature maps $X_o$ and $X_s$ obtained after the SFA and MFA modules, which serve as inputs to our dual-frame fusion module. 
%To combine semantic information of high-level feature maps and spatial information of the low-level feature maps, the proposed dual frame fusion module aggregates those feature maps. 
The proposed dual-frame fusion module combines semantic information of the high-level feature maps and spatial information of low-level feature maps. Instead of traditional up-sampling, inspired by \cite{zhang2022look}, we use Adaptive Feature Pooling for a more flexible approach. This offers an expanded receptive field, facilitating improved integration of both core and contextual semantics.

The high-level feature map is adaptively pooled to match the size of the low-level feature maps. These feature maps are then combined via pixel-wise summation and further processed through deformable convolutions. This offers better adaption to different object sizes, shapes, and other geometric deformations. The input has a total of four layers, and the aforementioned convolution and up-sampling process is iterated 2 times to have the final output. With the help of the above process, we obtained channel and spatial attention feature maps on single frames and temporal attention feature maps for multiple frames. These feature maps are aggregated to generate a fused feature map. 

\subsection{Detection Head}
Our detection head is similar to CenterNet~\cite{duan2019centernet}. We performed the computation of the fused object probability heatmap on the merged feature maps. However, the size and offset of the bounding boxes are generated from single-frame features. The loss function comprises three components: a fusion heatmap loss calculated with Focal Loss, and two regression losses (offset and size) computed with L1 Loss. The formulation of each loss is as follows. $L_{Z}$ is the unique fusion heatmap loss,
\begin{equation}
    L_{Z} = -\frac{1}{M} \sum_{ij} 
    \begin{cases} 
      (1-\hat{Q}_{ij})^\epsilon \log(\hat{Q}_{ij}) & \text{if } Q_{ij} = 1 \\
      (1-Q_{ij})^\zeta (\hat{Q}_{ij})^\epsilon \log(1-\hat{Q}_{ij}) & \text{otherwise}
   \end{cases},
\end{equation}
where  \( \hat{Q}_{ij} \) indicates the predicted heatmap value for each pixel, \( Q_{ij} = 1 \) signifies the pixel is the center of an object, and \( \epsilon \) and \( \zeta \) are the modified focal loss hyper-parameters. $L_{Y}$ represents the loss for heatmap offset,

\begin{equation}
    L_{Y} = \frac{1}{M} \sum_{q} | \hat{P}_{q} - T - \tilde{q} |,
\end{equation}
where \( \hat{P}_{q} \) is the predicted offset, \textit{T} is the position after downsampling, and \( \tilde{q} \) is the actual center point. $L_{X}$ calculates the loss for the size of the bounding box,

\begin{equation}
    L_{X} = \frac{1}{J} \sum_{j=1}^{J} | \hat{R}_{j} - R_{j} |,
\end{equation}
where \( \hat{R}_{j} \) is the predicted size and \( R_{j} \) is the ground truth size. The overall training objective is

\begin{equation}
    L_{total} = L_{Z} + \lambda_{dim} L_{X} + \lambda_{pos} L_{Y},
\end{equation}
where $\lambda_{dim}$ and $\lambda_{pos}$ are the adjusted hyper-parameters for the size and offset loss components, respectively.

\section{Experiments}
In this section, we assess the performance of our proposed method compared to SOTA methods and perform an ablation study. 
\subsection{Datasets and Evaluation Metrics}

\textbf{Datasets:} As our method relies on more than a frame, the evaluation requires the use of video datasets. Our selected evaluation domain focuses on traffic surveillance given its significant relevance to our research. We used datasets with videos, but some are not standard datasets for object detection. Nevertheless, they were used in previous work on video object detection. We chose: KITTI MOT (Multi-Object Tracking) \cite{geiger2012we} and Cityscapes \cite{elsayed2020revisiting}, both not used for object detection usually but provide videos, and UAVDT \cite{du2018unmanned} used for object detection in videos. Each of these datasets provides unique challenges and contains sequences at different viewpoints with different sizes of objects. As we are using non-standard datasets (KITTI, Cityscapes) for object detection, we needed to compute some results ourselves for competing SOTA methods for a fair comparison. However, this is not true for the UAVDT dataset, where we use the standard data training and test split.

\textbf{Evaluation Metrics:} We use Average Precision (AP) for multiple scales of objects and Mean Average Precision (mAP) across varying IoU thresholds and mAP50 and mAP75, respectively at 0.5 and 0.75 IoU thresholds, to evaluate detection accuracy. Intersection over Union (IoU) is used to evaluate bounding box precision on all datasets. 

 \subsection{Implementation Details}
For features extraction, we used HRNet \cite{wang2020deep}, and pre-trained it on the COCO dataset \cite{lin2014microsoft}, following the methodology described in \cite{duan2019centernet}. %COCO was chosen for its wide range of different objects, which helps the backbone to learn to recognize a variety of objects effectively.
 %We evaluated the proposed model by initializing the weight of our feature extractor on the COCO dataset\cite{lin2014microsoft}, 
Our global architecture follows CenterNet~\cite{duan2019centernet}. However, our training process is done in two steps. First, our backbone is fine-tuned on each dataset starting from the pre-trained weights on COCO. Then, the first and third layers of the backbone are frozen and the MFA, SFA, and dual-fusion modules as well as the network heads are trained. Training is conducted over 250 epochs utilizing the Adam optimizer, starting with a learning rate of $1 \times 10^{-4}$, which undergoes a decimation by a factor of 10 after the 130\textsuperscript{th} and 140\textsuperscript{th} epochs. To ensure training stability, we use gradient clipping. The same training protocol was used for the overall architecture as well as for all the base detectors to demonstrate the contribution of our approach.

\subsection{Results and Discussion}
%We have performed experiments on Cityscapes \cite{elsayed2020revisiting} , KITTI \cite{geiger2012we} and UAVDT \cite{du2018unmanned} datasets.\\
%\textbf{Results on the Cityscapes dataset.}
%The curve between metric and losses, as shown in fig \ref{fig:graph}, indicates how our proposed model increases its accuracy with an increase of every epoch, as expected. 
Comparisons with SOTA methods on the Cityscapes dataset are reported in Table \ref{tab:1}. 

They show that our attention-based fusion detector consistently outperforms the other SOTA detectors. There is a significant improvement in the detection results when using our STF model as compared to SOTA detectors. The improvement in detection results is due to our two attention modules and our dual-fusion module, all contributing positively to detecting objects better (especially small or occluded ones). In Table \ref{tab:1}, we also compare our model with the vanilla HRNet as we use a feature extractor based on the HRNet architecture. This allows us to examine our results in comparison to vanilla HRNet to observe the impact of our STF module on a similar backbone. This comparison demonstrates a gain in accuracy for all sizes of objects. Furthermore, we changed the backbone of Centernet \cite{duan2019centernet} to observe how HRNet affects its performance as it uses a detection similar as ours. It can be concluded from the results that using HRNet alone does not yield significant improvement. This is another demonstration that our method using a classification head similar to CenterNet performs better due to our SFA and MFA modules. By comparing our results with YOLOv5 and the recent YOLOX, our model shows improvement in terms of precision and accuracy, as well. Finally, we also perform better than PPNet which uses multiple frames. 

\begin{table*}[ht]
\centering
\caption{Comparison of our method with SOTA methods on the Cityscapes validation dataset. \textbf{Boldface} indicates best results.}
\label{tab:1}
\begin{tabular}{|c|c|c|c|c|c|c|c|c|}
\hline
\textbf{Method Type} & \textbf{Method} & \textbf{Backbone} & \textbf{$mAP$} & \textbf{$AP_{50}$} & \textbf{$AP_{75}$} & \textbf{$AP_S$} & \textbf{$AP_M$} & \textbf{$AP_L$} \\
\hline
\multirow{6}{*}{SOTA (Single Frame Detectors)} & RetinaNet \cite{li2023perspectivenet} & RetinaNet-50 & 92.8 & 94.1 & 93.2 & 43.1 & 58.2 & 94.0 \\
& Vanilla HRNet* & HRNet & 92.2 & 94.9 & 91.1 & 45.3 & 59.9 & 93.1 \\
& CenterNet \cite{li2023perspectivenet} & Hourglass-104 & 92.1 & 93.9 & 93.0 & 43.2 & 56.7 & 93.5 \\
& CenterNet* & Resnet-18 & 92.7 & 92.5 & 92.7 & 44.8 & 57.6 & 93.2 \\
& CenterNet* & HRNet & 92.8 & 93.3 & 93.1 & 45.9 & 58.7 & 93.4 \\
& YOLOv5 \cite{li2023perspectivenet} & CSPDarknet53 & 93.6 & 93.4 & 91.8 & 43.7 & 59.5 & 95.1 \\
& YOLOX\cite{li2023perspectivenet} & CSPDarknet53 & 93.9 & 94.9 & 92.7 & 44.8 & 61.5 & 96.7 \\
\hline
\multirow{2}{*}{SOTA (Two Frame Detectors)} & PPNet\cite{li2023perspectivenet} & Resnet-50 & 94.8 & 96.2 & 92.5 & 43.9 & 57.4 & 95.8 \\
& STF (Ours) & HRNet & \textbf{95.7} & \textbf{97.2} & \textbf{95.3} & \textbf{49.3} & \textbf{65.3} & \textbf{97.3} \\
\hline
\multicolumn{9}{l}{$^{\mathrm{*}}$Trained by ourselves.}
\end{tabular}
\end{table*}

Table \ref{tab:2} presents the results of the KITTI validation dataset. The conclusions are the same as for Cityscapes with similar improvements compared to baseline methods. By comparing it with other SOTA detectors, our proposed method outperforms them with improvements for all object size categories (Small, medium and large). Our method demonstrates an improvement in detection results when compared to SOTA single-frame and two-frame detectors.

Results on the UAVDT test dataset are reported in Table \ref{tab:3}. Our Spatio-Temporal Fusion (STF) module consistently outperforms the base detectors. As well, when compared to SOTA multi-frame detectors, such as FFAVOD and RN-VID that fuse features without attention, we can notice that although this helps compared to single-frame detectors, a more sophisticated fusion approach, like the one we propose is required to obtain even better results. 
\begin{table*}[ht]
\centering
\caption{Comparison of our method with SOTA methods on the KITTI MOT validation dataset. \textbf{Boldface} indicates the best result.}
\begin{tabular}{|c|c|c|c|c|c|c|c|c|}
\hline
\textbf{Method Type} & \textbf{Method} & \textbf{Backbone} & \textbf{$mAP$} & \textbf{$AP_{50}$} & \textbf{$AP_{75}$} & \textbf{$AP_S$} & \textbf{$AP_M$} & \textbf{$AP_L$}\\
\hline
\multirow{4}{*}{\shortstack{SOTA methods\\(Single Frame Detectors)}} & RetinaNet \cite{lorente2021scene}& RetinaNet-50 & 56.6 & - & - & 29.9 & 62.8 & 73.1 \\
& Vanilla HRNet* & HRNet & 79.1 & 81.6 & 70.0 & 48.3 & 65.2 & 74.3 \\
& CenterNet \cite{wang2023centernet} & Hourglass-104 & - & 85.3& - & - & - & - \\
& CenterNet* & Resnet-18 & 80.5 & 83.4 & 74.5 & 50.2 & 66.8 & 78.7\\
& CenterNet* & HRNet & 81.7 & 83.3 & 74.1 & 50.0 & 66.8 & 77.4 \\
& YOLOv5* & CSPDarknet53 & 84.3 & 86.8 & 76.3 & 52.9 & 70.4 & 83.5 \\
& YOLOX* & CSPDarknet53 & 85.9 & 87.7 & 79.8 & 53.8 & 71.7 & 84.9 \\
\hline
\multirow{4}{*}{\shortstack{SOTA methods\\(Two Frame Detectors)}} & Mf-SSD \cite{broad2018recurrent} & SqueezeNet & 83.0 & - & - & - & - & - \\
& MFCN \cite{liu2020video} & ResNet101 & 84.6 & - & - & - & - & - \\
& PPNet \cite{li2023perspectivenet} & ResNet50 & 86.2 & - & - & - & - & - \\
& STF (Ours) & HRNet & \textbf{88.7}& \textbf{90.0}& \textbf{82.9}& \textbf{57.1}& \textbf{74.6}& \textbf{88.1} \\
\hline
\multicolumn{9}{l}{$^{\mathrm{*}}$Trained by ourselves.}
\end{tabular}
\label{tab:2}
\end{table*}

\begin{table*}[ht]
\centering
\caption{Comparison of UAVDT test dataset with different methods. \textbf{Boldface} indicates the best result.}
\begin{tabular}{|c|c|c|c|c|c|c|c|c|}
\hline
\textbf{Method Type} & \textbf{Method} & \textbf{Backbone} & \textbf{$mAP$} & \textbf{$AP_{50}$} & \textbf{$AP_{75}$} & \textbf{$AP_S$} & \textbf{$AP_M$} & \textbf{$AP_L$}\\
\hline
\multirow{5}{*}{\shortstack{SOTA methods\\(Single Frame Detectors)}} & RetinaNet \cite{zhang2021multi}& RetinaNet-50 & 16.2& 34.0 & 13.7&8.8& 30.1 &23.8  \\
& CenterNet \cite{liao2021unsupervised}  & Hourglass-104 & 16.4 & 29.7 & 16.6 & 12.2 & 25.1 & 11.3\\
& YOLOv5 \cite{liao2021unsupervised} & CSPDarknet53 & 18.0 & 33.5 & 17.2 & 11.0 & 29.6 & 37.5 \\
& YOLOX\cite{xu2022stn} & CSPDarknet53  & 26.0 & 43.3 &21.4 & - & - & - \\
& SpotNet \cite{perreault2020spotnet} & U-Net   & 53.4 & -  &- & - & - & -\\
\hline
\multirow{6}{*}{\shortstack{SOTA methods\\(Two Frame Detectors)}} & STDnet-ST \cite{wu2021deep} &  RCN  & 13.3 & 36.4  &- & - & - & - \\
& AdNet-MS \cite{zhang2021multi} &  Darknet53  & 13.3 & 43.5  &18.3 & 12.1 & 37.9 & 27.9 \\
& RN-VID \cite{perreault2020rn} &    & 39.4 & -  &- & - & - & -\\
& FFAVOD-SpotNet \cite{perreault2021ffavod} &    & 53.8 & -  &- & - & - & -\\
& FFAVOD-CenterNet \cite{perreault2021ffavod} &    & 52.1 & -  &- & - & - & -\\
& STF (Ours) & HRNet & \textbf{58.4}& \textbf{79.5} & \textbf{46.3}& \textbf{35.8}& \textbf{59.4}& \textbf{61.9}\\
\hline
\end{tabular}
\label{tab:3}
\end{table*}

\begin{table}[ht]
\centering
\caption{Ablation study on the MFA, SFA, and Dual-fusion modules}
\small
\begin{tabular}{|c|c|}
\hline
\textbf{Configuration} & \textbf{mAP (\%)} \\
\hline
Baseline (HRNet+CenterNet head) & 92.10 \\
Baseline + SFA & 93.50 \\
Baseline + MFA & 94.91 \\
Baseline (MFA+SFA) & \textbf{95.73} \\
\hline
\end{tabular}
\label{tab:comparison}
\end{table}

\subsection{Ablation Study}
An ablation study was performed to evaluate the contribution of the different parts of the proposed method:  the multi-frame attention module, single-frame attention module as well as the single-frame and multi-frame attention with the dual-frame fusion module, and show the effect of each component in Table \ref{tab:comparison}. We find that the method with the MFA module or the method with the SFA module both detect better than the baseline method (HRNet + CenterNet head). We also observe that the proposed method (STF) with both two modules and dual-frame fusion performs the best. According to the proposed STF module, our MFA module plays a crucial role in combining features from two frames. Similarly, the SFA module aims to improve the accuracy of detection within a single frame. This is achieved through the combination of single-frame channel and spatial attention, which effectively suppresses false positive detection in background regions. In our observations, we noted that each module independently contributes to performance enhancement. Moreover, a synergistic effect is observed when both modules are combined, leading to a more significant improvement in results. Therefore, for better efficiency and accuracy, our proposed model demonstrates superior results as compared to other configurations.

To illustrate the specific contributions of our proposed dual-frame fusion method, we also conducted an ablation study on it. We aimed to understand the individual impact of different fusion strategies on the overall performance of our model. For that, we use different strategies of combining two frames, i.e. concatenation, median, mean, and max fusion. In all cases, that decreased the performance by a large margin as shown in Table \ref{tab:fusion}. We attribute this to the misalignment of features across frames, necessitating a more intricate operation for aggregating these features. Admittedly, our model requires additional parameters to effectively learn the optimal combination of feature maps. However, as indicated in Table \ref{tab:fusion}, our findings strongly support the benefit of our dual-frame fusion method in integrating feature maps.

\begin{table}[ht]
    \centering
    \caption{Ablation study of the different fusion strategies on Cityscapes dataset.}
    \small
    \begin{tabular}{|c|c|}
        \hline
        \textbf{Fusion Methods} & \textbf{mAP}\\
        \hline
        \text{Concatenation} & 88.60\\
        \hline
        \text{Median} & 91.50 \\
        \hline
        \text{Mean} & 91.70 \\
        \hline
        \text{Max} & 91.89 \\
        \hline
        \text{Dual-frame fusion (Ours)} & \textbf{95.73}\\
        \hline
    \end{tabular}
    \label{tab:fusion}
\end{table}

%\begin{figure*}[ht]
%\centering
%\includegraphics[width=\textwidth]{images/output.png} 
%\caption{Qualitative results for Cityscapes, KITTI, and UAVDT datasets, respectively.}
%\label{fig:output}
%\end{figure*}

\section{Conclusion}
In this work, we designed a spatio-temporal fusion module as a new approach for multi-frame object detection. Specifically, we identified the ineffectiveness and inadequacy issues present in single-frame object detectors. Then, we proposed to solve these problems using multi-frame and single-frame attention modules, as well as a dual-frame fusion module to improve object representation. Our results show that by exploiting sequential frames, we can improve the efficiency and accuracy of detection under challenging conditions.

%This dual-frame approach, with its unique combination of temporal, channel, and spatial attention, represents a significant advancement in feature map processing for sequential analysis, to improve the object detection.


\section*{Acknowledgement}
We acknowledge the support of the Natural Sciences and
Engineering Research Council of Canada (NSERC), [funding reference number RGPIN-2020-04633].


\bibliographystyle{IEEEtran}
%\bibliography{Refrences}

%\bibliographystyle{IEEEtran}

\end{document}